\documentclass{article}
\usepackage{wrapfig}
\usepackage{xcolor}
\usepackage{enumitem}

\usepackage[preprint]{corl_2026} 
\usepackage{booktabs}
\usepackage{graphicx}
\usepackage{float}
\usepackage{amssymb}
\usepackage{bbm}
\usepackage{multirow}
\usepackage{pifont}
\usepackage{amsmath}
\usepackage[capitalize]{cleveref}

\newcommand{\method}{\textcolor[HTML]{002676}{\textsc{Do as I Do}}}

\newcommand{\guide}[1]{\textcolor[HTML]{1F5FAA}{#1}}
\newcommand{\refz}[1]{\textcolor[HTML]{D97A1C}{#1}}
\newcommand{\euler}[1]{\textcolor[HTML]{7A4FB5}{#1}}

\newcommand\blfootnote[1]{%
  \begingroup
  \renewcommand\thefootnote{}\footnote{#1}%
  \addtocounter{footnote}{-1}%
  \endgroup
}

\title{Do as I Do: Dexterous Manipulation Data\\ from Everyday Human Videos}

\author{
\normalfont Bhawna Paliwal$^{*}$ \quad Haritheja Etukuru$^{*}$ \quad William Liang$^{*}$ \\
Pieter Abbeel \quad Nur Muhammad ``Mahi'' Shafiullah \quad Jitendra Malik \\[0.5em]
\textbf{UC Berkeley} \\[1.2em]
\url{https://do-as-i-do.com}
}

\begin{document}
\maketitle
\blfootnote{$^{*}$Denotes equal contribution. Correspondence to: \texttt{bhawna\_paliwal@berkeley.edu}.}


\vspace{-4em}

\begin{abstract}
How can we scalably generate data for robotic manipulation, especially on human-like platforms such as dexterous multi-fingered hands?
{Learning from human videos} has recently emerged as a likely answer to this question.
However, difficulties in estimating hand-object interaction and crossing the human-to-robot embodiment gap have hindered the adoption of abundant monocular RGB-only human videos as the \textit{primary} source of robot manipulation data.
In this work, we present \method{}, an algorithm to reconstruct and retarget monocular RGB human videos to multi-fingered dexterous robotic hands.
\method{} reconstructs hand-object interactions from various egocentric and exocentric in-the-wild video sources.
The algorithm then retargets these hand-object interaction estimates into a sequence of actions executable in the real world, yielding robot-complete manipulation data from disparate human videos.
Overall, \method{} outperforms previous state of the art in estimating hand-object interactions and extracting dexterous manipulation trajectories from RGB videos, as we show in experiments on datasets with ground truths and on a dataset of video clips collected online.
Our experiments enable us to propose an efficacy playbook for practitioners collecting human data for manipulation. 

\end{abstract}

\begin{figure}[H]
\centering
\includegraphics[width=\columnwidth]{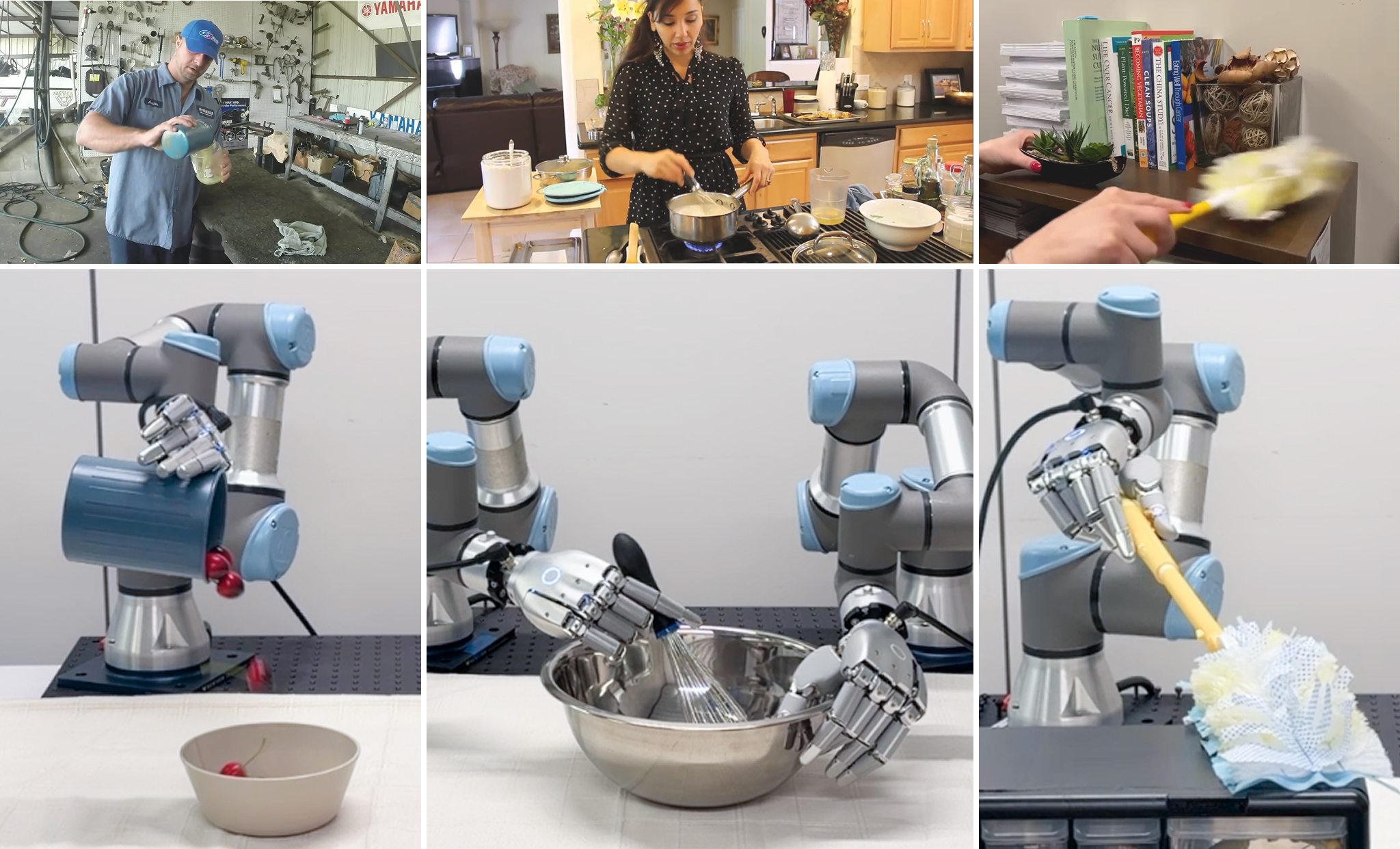}
\vspace{-1.5em}
\caption{We introduce \method{}, an algorithm that takes in-the-wild monocular RGB videos of hand-object interaction (top) and generates dexterous hand manipulation data (bottom).}
\label{fig:title}
\end{figure}


\section{Introduction}
\label{sec:intro}

Data is a critical component of any learning algorithm, including robot learning. 
For a novice intelligent actor, be it a child, an apprentice, or a robot, however, available data for any task is mostly \textit{observational}: experienced not via \textit{doing} but via \textit{watching} experts do it. This gap in watching and doing is easily closed by children~\cite{meltzoff_imitation_1977, meltzoff_infant_1988}, but presents an insurmountable barrier for today's robots, which rely primarily on \textit{experiential} data collected via real-world teleoperation or simulated exploration. 
Consider the challenges of generating experiential data for dexterous behaviors as shown in~\cref{fig:title}.
Teleoperation is bottlenecked by operator expertise, cost of operation, and mechanical transparency of the teleoperation rig.
Exploration in simulation is similarly bottlenecked by complexities in designing diverse environments and reward functions.
The natural question we ask in this work is how to close this gap by converting the accessible, observational data from humans into experiential data for robots.

Investigations into inferring robot actions from watching humans, also known as ``Do as I do'', are almost as old as the field of artificial intelligence, seen notably in the 1970 MIT ``copy-demo''~\cite{mit_copy_demo}, ``robot instruction'' by direct human demonstration~\citep{554349} and ``Do as I do'' with action synthesis and retargeting~\citep{Efros03}.
The two primary algorithmic challenges are {recognition} or {reconstruction} of the human behavior, and retargeting of this behavior on a (possibly vastly different) robotic embodiment.
Assumptions on the behaviors, i.e.\ pick-and-place only, and on the objects, i.e.\ 3D-scanned only, have made inroads into this challenge possible~\cite{qin2022dexmv,mu2026deximit}.
Over time, further assumptions on the data modality, such as availability of depth, 3D, or hand keypoints in data collected with specialized hardware, have made more advances in ``Do as I do'' possible~\cite{guzey2025dexterity,liu2025egozero}.
However, most of our observational data today is stored as monocular RGB videos of humans, and therefore algorithms that can address this general case stand to deliver the largest increase in robotic data.

Here, two recent advances in adjacent fields provide us with a novel approach to this problem. First, 3D computer vision models today can take 2D RGB images and reconstruct depth~\citep{wang2025moge2accuratemonoculargeometry}, objects~\citep{sam3dteam2025sam3d3dfyimages}, and hands~\citep{pavlakos2024reconstructing} in 3D from purely monocular RGB videos, enabling 4D hand-object estimation.
Simultaneously, GPU-parallel physical simulators such as Mujoco Warp~\cite{mujoco_warp} and Isaac~\cite{nvidia_isaac_sim} enable fast sampling-based optimization algorithms~\cite{pan2026spiderscalablephysicsinformeddexterous} that we can use to infer robotic dexterous hand actions in minutes from 4D hand-object states.
Note that both of these approaches make minimal assumptions on the problem structure, such as the observed behavior or target objects.
Inspired by such advances, in this work we present \method{}, aiming to bridge this gap between observational and experiential data for dexterous manipulation.~\method{} recognizes and retargets dexterous human actions from everyday monocular RGB videos and produces multi-fingered robotic hand and arm actions performing the same task on the environment.
More importantly, we are able to do so without making limiting assumptions on the displayed behavior (e.g.\ no grasping priors) or object classes, supporting arbitrary rigid bodies.
Concretely, the contributions of this work are as follows:
\begin{enumerate}[leftmargin=*]
  \item We introduce~\method{}, a two-step algorithm to reconstruct and retarget behaviors from monocular RGB videos to multi-fingered dexterous hands.
  \item Our hand-object reconstruction process outperforms SOTA on relevant metrics and handles diverse videos — ego- or exo-centric, ranging from in-the-wild internet clips to outputs of generative video models.
  \item Our retargeting process improves upon existing scalable dynamics-aware retargeting techniques by introducing novel components that robustify the noisy reconstructed reference trajectories.
  \item Our robot data is playable on a dexterous robot hand and arm, completing, to the best of our knowledge, the first pipeline that can go from an internet video to real dexterous hand rollouts.
\end{enumerate}

\section{Related Work}
\label{sec:related-work}

\begin{table}[t]
    \centering
    \caption{\textbf{Related Work.} We summarize methodology and data sources used by prior work, in rough order of difficulty. Self refers to data collected by the authors. 
    }
    \label{tab:related-work}
    \resizebox{\columnwidth}{!}{%
    \begin{tabular}{l cc cccc}
        \toprule
        & \multicolumn{2}{c}{Method} & \multicolumn{4}{c}{Data Sources} \\
        \cmidrule(lr){2-3} \cmidrule(lr){4-7}
        & Reconstruction & Retargeting & Self & Gen. & Ego. & Internet \\
        \midrule
        H2Sim2Robot~\citep{lum2025crossinghumanrobotembodimentgap} & LiDAR Scan + FPose~\citep{wen2024foundationposeunified6dpose} & RL & $\checkmark$ & & & \\
        VideoManip~\citep{chen2026videomanip} & MeshyAI~\citep{meshyai2025} + FPose & DRO~\citep{wei2025mathcaldrograspunifiedrepresentation} + DP3~\citep{Ze2024DP3} & $\checkmark$ & & $\checkmark$ & \\
        DexMan~\citep{hsieh2025dexman} & TRELLIS~\citep{xiang2024structured} + FPose + SpaTrack~\citep{xiao2025spatialtrackerv23dpointtracking} & RL & $\checkmark$ & $\checkmark$ & & \\
        DexImit~\citep{mu2026deximit} & SAM\,3D~\citep{sam3dteam2025sam3d3dfyimages} + FPose++~\citep{yan2025foundationposeplusplus} &  Motion planning & $\checkmark$ & $\checkmark$ & & \\
        \midrule
        \textbf{Ours} & SAM\,3D + Guided Diffusion & Sampling-based opt. & $\checkmark$ & $\checkmark$ & $\checkmark$ & $\checkmark$ \\
        \bottomrule
    \end{tabular}%
    }
\end{table}

\textbf{Dexterous Manipulation from Human Videos.} Many past works explore ways to leverage semantics and motions from human videos. Some extract priors via pretraining, either as visual representations~\citep{ma2023vipuniversalvisualreward,ma2023livlanguageimagerepresentationsrewards,nair2022r3muniversalvisualrepresentation}, dexterous policies~\citep{shaw2022videodexlearningdexterityinternet,zheng2026egoscalescalingdexterousmanipulation,yang2025egovlalearningvisionlanguageactionmodels,luo2025beingh0visionlanguageactionpretraininglargescale,punamiya2026egoverseegocentrichumandataset}, or forward dynamics models~\citep{goswami2026worldmodelslearningdexterous,gao2026dreamdojogeneralistrobotworld}. Other approaches compute more structured priors, such as affordances~\citep{shi2025zeromimicdistillingroboticmanipulation,agarwal2023dexterousfunctionalgrasping}, flows~\citep{bharadhwaj2024track2actpredictingpointtracks,wang2023mimicplaylonghorizonimitationlearning}, and 3D reconstructions for retargeting~\citep{qin2022dexmv,singh2024handobjectinteractionpretrainingvideos,qin2023handmultiplehandsimitation,li2024okamiteachinghumanoidrobots,lum2025crossinghumanrobotembodimentgap,mu2026deximit,hsieh2025dexman,chen2026videomanip}. Our work falls in the final category and pushes the frontier in utilizing diverse in-the-wild data sources, as shown in~\cref{tab:related-work}.

\textbf{Hand-Object Reconstruction.} Reconstructing 4D hand-object interaction from monocular RGB video decomposes along three complementary axes: hand pose estimation, object shape and pose estimation, and their joint modeling. The structured shape and motion of the human hand~\citep{mano} has 
enabled hand tracking models~\citep{pavlakos2024reconstructing, potamias2024wilor, zhang2025hawor} robust to motion blur, occlusion, and low resolution. 
\emph{Object reconstruction and tracking} under the same in-the-wild noisy conditions remain substantially harder because of diversity in everyday objects. Recent progress has come from (1) image-conditioned 3D generative foundation models~\citep{liu2023one2345,xiang2024structured,sam3dteam2025sam3d3dfyimages} with robust priors over shape and pose, which can handle occlusions and lower resolution, and (2) model-based 6-DoF trackers~\citep{wen2024foundationposeunified6dpose,lee2025any6d} that estimate object pose given a known or jointly reconstructed mesh. However, these methods were largely validated on clean lab videos and struggle on in-the-wild noisy videos, as we show below.
Finally, \emph{joint hand-object reconstruction} methods reason about both hand and object signals simultaneously, and have shown progress on lab data~\citep{hasson2019learning} as well as in-the-wild videos~\citep{ye2022s,prakash20243d,wu2024reconstructing}. 
Recent video methods remain narrow in scope: some are egocentric-only~\citep{ye2025whole}, while category-conditioned approaches~\citep{ye2023vhoi,ye2023ghop} assume closed object taxonomies.
In contrast to these joint approaches, \method{} adopts a \emph{modular decomposition}: HaWoR for hand tracking, SAM3D for single-image object meshing, and our SAM3D-based tracking (Sec.~\ref{subsec:reconstruction}) for object pose evolution. 

\textbf{Human-to-Robot Retargeting.} Dexterous retargeting algorithms map human hand poses onto robot embodiments with vastly different geometries (e.g., finger link lengths and articulations). \emph{Kinematic retargeting} approaches do so by solving geometric, task-space, or joint-space optimizations~\citep{Zakka_Mink_Python_inverse_2026,kim2025pyrokimodulartoolkitrobot,qin2024anyteleopgeneralvisionbaseddexterous,yin2025geometricretargetingprincipledultrafast}, but operate solely on the robot configuration and do not account for forces between hand and object, often causing penetration, fingertip sliding, and grasp instability. To address this, \emph{dynamics-aware retargeting} frameworks optimize physically-simulated hand-object trajectories, generally via reinforcement learning to track a reference-based reward~\citep{li2025maniptransefficientdexterousbimanual,mandi2025dexmachinafunctionalretargetingbimanual,lum2025crossinghumanrobotembodimentgap,xu2025dexplorescalableneuralcontrol} or sampling-based optimization~\citep{yang2026physicsdrivendatagenerationcontactrich,si2025exostartefficientlearningdexterous,pan2026spiderscalablephysicsinformeddexterous}. However, a key assumption shared by most prior retargeting approaches is availability of clean references with ground-truth hand-object poses from, e.g., MoCap. In contrast, we tackle the more difficult setting of noisier reconstructed references, with potential temporal discontinuities and severe hand-object misalignments.


\section{Method}
\label{sec:method}

\begin{figure}
\centering
\includegraphics[width=\columnwidth]{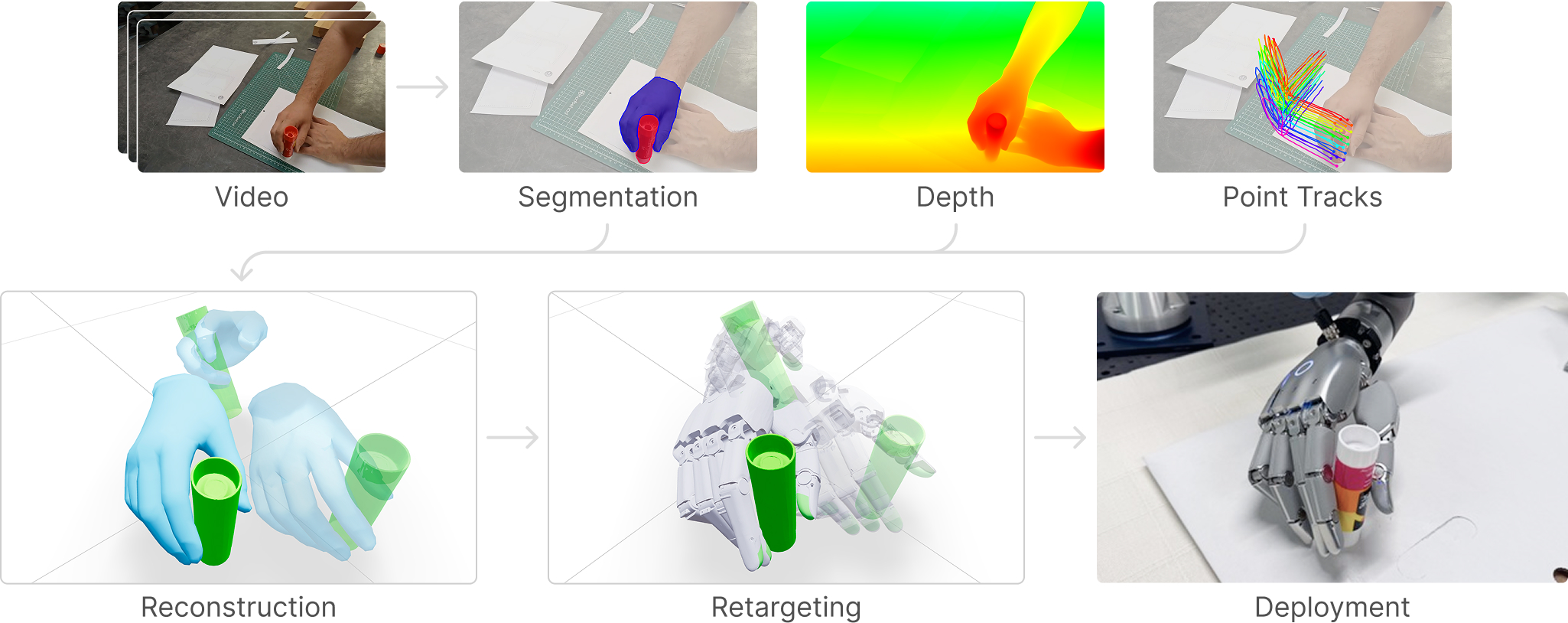}
\vspace{-1em}
\caption{\textbf{Method Overview.} Our method leverages vision foundation models to reconstruct the hand and object, and retargets them onto the robot via sampling-based optimization in simulation.}\label{fig:method}
\vspace{-0.5em}
\end{figure}
\method{} consists of two parts, shown in~\cref{fig:method}.
First, we reconstruct the 3D hand and object, and track them through time (\cref{subsec:reconstruction}). Then, we retarget the reconstructions onto the robot embodiment, producing dynamically-feasible trajectories that are effective in the real world (\cref{subsec:retargeting}).

\subsection{Reconstruction}
\label{subsec:reconstruction}

Recognizing and reconstructing hand-object interactions from in-the-wild videos can be decomposed into two components: (1) tracking the human hand, and (2) determining object shape and tracking its pose.
Critically, these capabilities need to be robust to diverse visual conditions found in noisy internet videos. We find that existing models such as HawoR~\citep{zhang2025hawor} satisfy this criterion and can directly be used for our hand tracking with reasonable performance.
For open-ended rigid objects with hand occlusions, however, prior works~\citep{wen2024foundationposeunified6dpose, lee2025any6d} tend to lose pose lock, drift, or fail to re-acquire the object once visual evidence degrades as shown in~\ref{fig:tracking-comparison}.
Therefore, we develop our own tracking method based on 3D generative foundation models trained with occlusions, namely SAM\,3D~\citep{sam3dteam2025sam3d3dfyimages}.
Once tracked, the 3D hand, object, and camera are composed into a consistent near metric-space.
We run three processing steps: (a) hand and object segmentation using SAM\,3~\citep{carion2026sam3segmentconcepts}, (b) depth and camera intrinsics estimation with MoGe~\citep{wang2025moge2accuratemonoculargeometry}, and (c) object 3D mesh generation using SAM\,3D~\citep{sam3dteam2025sam3d3dfyimages}. 

\textbf{Object Tracking via Guided Diffusion.} 
We repurpose SAM\,3D~\citep{sam3dteam2025sam3d3dfyimages}, an image-to-3D generative model robust to occlusions and low resolution, into a video object tracker.
It learns the joint distribution over shape and pose $p_\theta(x^{s}, x^{p} \mid c)$ given a single 2D image and object mask. 
As a result, it produces a \emph{different} mesh each frame and a pose sequence with no temporal coherence if applied independently on frames.
Our key observation is that shape and pose share the same latent space in the learned joint distribution; we can therefore fix the shape and obtain an updated pose entirely at inference time for each frame. Specifically, we fix the shape $\bar x^{s}$ at an \emph{anchor} frame and, given the pose $x^{p}_{k-1}$ from the previous frame, predict the pose $x^{p}_{k}$ at frame $k$. Tracking thus reduces to drawing from $p_\theta(x^{p}_k \mid x^{s}_k = \bar x^{s},\, c_k)$ biased toward $x^{p}_{k-1}$. 
Marginalizing over all possible poses is infeasible in the $6$-DoF continuous pose space with the large generative backbone of SAM\,3D. 
We instead exploit the flow matching inference itself: a sample is produced by integrating the ODE $\dot x = v_\theta(x_t, t, c)$ from $x_0 \!\sim\! \mathcal{N}(0, \mathbf{I})$ along the linear path $x_t = (1-t)\, x_0 + t\, x_1$. 
In a flow model, the forward-noised target is simply its \emph{interpolant} along the model's own probability path~\citep{lugmayr2022repaintinpaintingusingdenoising, song2021scorebasedgenerativemodelingstochastic}. At each Euler step, we take the model's \euler{free Euler update} of each block and blend it toward \refz{target interpolants}, nudging towards canonical shape $\bar x^{s}$ for the shape block, and previous-frame pose $x^{p}_{k-1}$ for the pose block:
\begin{equation}
x^{s}_{t} = \underbrace{(1-\guide{\alpha_{s}})(\euler{x^{s}_{t-\Delta} + \Delta v^{s}_{\theta}})}_{\text{denoising}} + \underbrace{\guide{\alpha_{s}}\, \refz{z^{s}_{\mathrm{ref}}(t)}}_{\text{blending}},
\quad
x^{p}_{t} = \underbrace{(1-\guide{\alpha_{p}})(\euler{x^{p}_{t-\Delta} + \Delta v^{p}_{\theta}})}_{\text{denoising}} + \underbrace{\guide{\alpha_{p}}\, \refz{z^{p}_{\mathrm{ref}}(t)}}_{\text{blending}}
\label{eq:interp}
\end{equation}
where $\guide{\alpha_{s}, \alpha_{p}} \in [0,1]$ are \guide{guidance strength} parameters;
$\refz{z^{s}_{\mathrm{ref}}(t)} = (1-t)\,\epsilon^{s} + t\,\bar x^{s}$ and
$\refz{z^{p}_{\mathrm{ref}}(t)} = (1-t)\,\epsilon^{p} + t\,x^{p}_{k-1}$ are \refz{target interpolants}; 
$\epsilon^{s}, \epsilon^{p}$ are the blocks' initial noise and 
$v^{s}_{\theta}, v^{p}_{\theta}$ are the shape and pose components of the velocity $v_\theta(x_{t-\Delta}, t-\Delta, c)$.

\textbf{Adaptive Guidance Parameters.}
As we focus on rigid objects, any fixed shape guidance $\guide{\alpha_{s}}\in[0.9, 1]$ works well. Instead of defining a fixed pose guidance $\guide{\alpha_{p}}$, which may cause over-rigidity or spurious flips, we derive it from the data using rotational velocity of the object estimated from 2D point tracks~\citep{doersch2024bootstapbootstrappedtrainingtrackinganypoint}. 
This adds one offline tracking pass per video but noticeably improves pose tracking as shown in Appendix.

\textbf{Sampling Per-frame Poses.} As explained above, the guided pose sampling in Eq.~\ref{eq:interp} is stochastic. So, at each frame $k$, our algorithm draws $N$ candidates $\{x^{p}_{k,i}\}_{i=1}^{N}$ that share the fixed shape $\bar x^{s}$ and needs to pick one of the samples per frame. The principled choice is to rank candidates by the model's own conditional log-density over poses given a shape. 
However, this process takes orders of magnitude more compute over generation itself and becomes prohibitive at video scale. We hence sample and cluster poses under a weighted $\mathrm{SE}(3)$ distance. Empirically, confident samples concentrate on the same pose mode while estimator noise scatters across $\mathrm{SE}(3)$, so consensus filtering and mask-IoU recovers the mode-best pose without ever re-invoking the diffusion backbone.

\textbf{Hand-Object Alignment.} 
After we independently reconustruct hand and object at possibly different scales, we need to align them.
We treat the hand reconstruction scale as ground-truth and scale the object translation $s$ to be aligned with the hand. 
We compute hand and object centroids: $\mathbf{c}^{\mathrm{M}}_{\mathrm{hand}}$, $\mathbf{c}^{\mathrm{M}}_{\mathrm{obj}}$ in object scale, and centroid of the \textit{visible} portion of the hand mesh $\mathbf{c}^{\mathrm{H}}_{\mathrm{hand}}$ in hand mesh scale (near metric). Now, given the scaling ratio from the centroid $z$ values, $k = z^{\mathrm{H}}_{\mathrm{hand}} /z^{\mathrm{M}}_{\mathrm{hand}}$, we optimize for the target object position $\mathbf{obj}_{\mathrm{target}} = \mathbf{c}^{\mathrm{H}}_{\mathrm{hand}} + k \,\bigl(\mathbf{c}^{\mathrm{M}}_{\mathrm{obj}} - \mathbf{c}^{\mathrm{M}}_{\mathrm{hand}}\bigr)$, where the per-frame translation scale is solved by least squares.
Finally, we align the trajectory with gravity using GeoCalib~\cite{veicht2024geocalib}.

\begin{figure}
\centering
\includegraphics[width=\columnwidth]{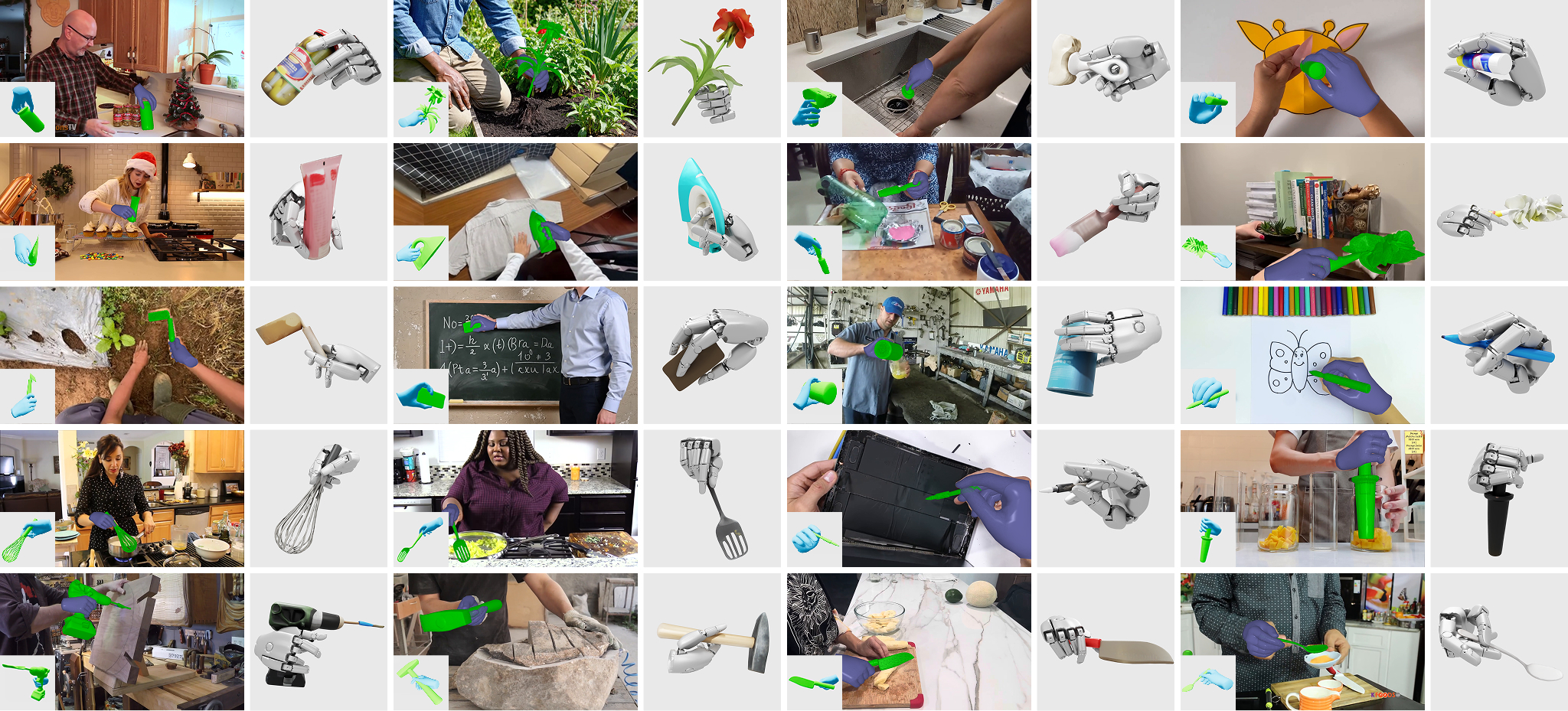}
\vspace{-1em}
\caption{\textbf{Verbs and Objects.} We visualize 20 distinct actions from our pipeline: placing, picking, scrubbing, spreading, squeezing, ironing, painting, dusting, digging, erasing, pouring, writing, whisking, stirring, poking, tamping, drilling, hammering, cutting, and basting.}
\label{fig:data}
\vspace{-0.5em}
\end{figure}

\subsection{Retargeting}
\label{subsec:retargeting}

Next, we aim to reproduce the reconstructed hand-object trajectory on a robot hand. However, this reference is incomplete: human and robot morphologies differ, and contact information and forces are absent from the kinematic signal. Prior works address this with kinematic solvers~\cite{hsieh2025dexman} or robotic heuristics~\cite{chen2026videomanip,mu2026deximit}, but they do not ensure physical plausibility or lose general-purpose expressiveness.

\begin{figure}
\centering
\includegraphics[width=\columnwidth]{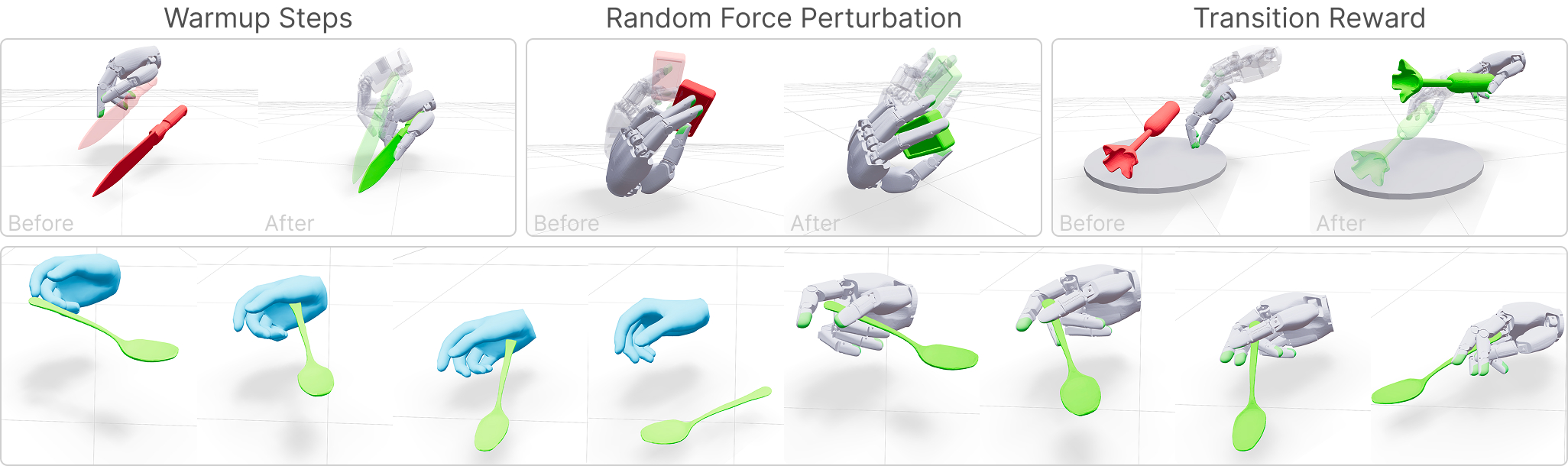}
\vspace{-1.5em}
\caption{\textbf{Retargeting.} Our method succeeds in common failure modes (top) and excels at handling noisy references (bottom), despite, e.g., incorrect depth estimation causing poor alignment.}\label{fig:method-retargeting}
\end{figure}

\method{} instead performs dynamics-aware retargeting, which follows the reference while ensuring realism within physics simulation. Building on the framework from~\citet{pan2026spiderscalablephysicsinformeddexterous}, we perform an MPPI-style sampling-based optimization with a kernel annealed across both iterations and the prediction horizon, which shifts from broad exploration to local refinement. To enable retargeting for noisy reconstructed references, we further highlight several novel innovations, as shown in~\cref{fig:method-retargeting}.

\textbf{Warmup Steps.} Two issues lie in the initial trajectory horizon $H$: (1) a noisy first frame may initialize the hand and object in a state that's impossible to recover from (e.g., if object is not grasped), and (2) annealed sampling does not fully explore these $H$ steps since they appear only at the start of the rollout horizon. Thus, we introduce additional $H$ warmup steps prepended to the reference. During warmup, the object is held in place (e.g., in mid-air) while the robot hand is free to move; afterwards, the weld is dropped and simulation proceeds as normal. This allows the robot to adjust its pose before tracking the reference (e.g., to avoid dropping the object in~\cref{fig:method-retargeting}), and naturally guides the optimizer in maximizing its tracking objective. Crucially, this warmup design does not assume any grasp sampling or heuristics, and simply utilizes the pre-existing core optimization procedure.

\textbf{Random Force Perturbation.} Second, the rollout horizon may trap optimization in local minima, with unstable object interactions that track briefly but cannot recover (e.g., balancing on fingertips in~\cref{fig:method-retargeting}). To address this, we assert that interactions should be robust against minor disturbances: drawing inspiration from sim-to-real~\citep{openai2019solvingrubikscuberobot,rudin2022learningwalkminutesusing}, we introduce random forces to sample rollouts, thus encouraging controls robust to such perturbations. Importantly, this solution is general-purpose and does not assume high-fidelity references, unlike alternatives (e.g., contact guidance~\citep{pan2026spiderscalablephysicsinformeddexterous}).

\textbf{Transition Reward.} Third, object transitions between ``rest" and ``in-hand" mark critical inflection points in the trajectory, but with noisy references, tracking reward alone is too imprecise and soft to encourage the step-function interaction induced by these transitions (e.g., failed pickup in~\cref{fig:method-retargeting}). Thus, we add a constant penalty term for failed transitions: (1) lack of object-floor contact during resting reference timesteps and (2) lack of hand-object contact during in-hand reference timesteps. We define reference timestep stages by measuring reference hand-object distance under threshold $\epsilon$.

\section{Experiments}
\label{sec:experiments}
\subsection{Experimental Setup}

We evaluate each step of our framework on standard benchmarks for their respective tasks. First, for hand-object reconstruction, we follow the setup from prior work~\citep{wu2024reconstructing, ye2023vhoi, ye2023ghop}: we evaluate on DexYCB~\cite{chao2021dexycb} and HOI4D~\cite{liu2022hoi4d} datasets with 160 and 12 annotated videos respectively, and isolate object-level performance by supplying ground-truth hands and measuring object reconstruction and tracking quality. 
We compare against two groups of baselines: (1) joint hand-object reconstructions including both image-based~\citep{hasson2019learning, ye2022s, prakash20243d} and video-based~\citep{wu2024reconstructing, ye2023ghop} approaches, and (2) object trackers~\citep{wen2024foundationposeunified6dpose, lee2025any6d}, where we replace our object tracking approach with these baselines keeping every other component fixed in our pipeline.
To additionally assess performance on a distribution closer to everyday videos, we collect a benchmark of 150 videos drawn from in-the-wild internet videos, egocentric datasets, and generated videos. Since ground-truth object poses are unavailable for these videos, we evaluate via human preference, asking 3 volunteers per video to compare object poses from our SAM\,3D-based tracking method against those from the current state of the art. Details of each of the baselines and human evaluation setup have been provided in the Appendix.

Finally, for retargeting, we evaluate on our in-the-wild reconstruction dataset of 655 reconstructed references, as well as OakInk2~\citep{zhan2024oakink2datasetbimanualhandsobject}, a large MoCap dataset with 1,352 clean bimanual human-object task trajectories. We compare against SPIDER~\citep{pan2026spiderscalablephysicsinformeddexterous}, the state-of-the-art for dexterous retargeting and the only prior method designed for this scale; SPIDER serves as the Annealed Sampling baseline, and we progressively introduce our three components to assess their contributions. Following the recent literature~\citep{pan2026spiderscalablephysicsinformeddexterous,li2025maniptransefficientdexterousbimanual,mandi2025dexmachinafunctionalretargetingbimanual}, we evaluate successful trajectories as those with mean position error $E_{\text{pos}} < 0.1$ m and mean rotation error $E_{\text{rot}} < 0.5$ rad. Technical details are in the Appendix.

Across all tasks, we use the 22-DoF Sharpa Wave hand. Real-world deployment results shown here are on a bimanual setup with Sharpa Wave hands and UR3e arms, both commanded at 50 Hz.

\subsection{Reconstruction Results}
\begin{figure}
\centering
\includegraphics[width=\columnwidth]{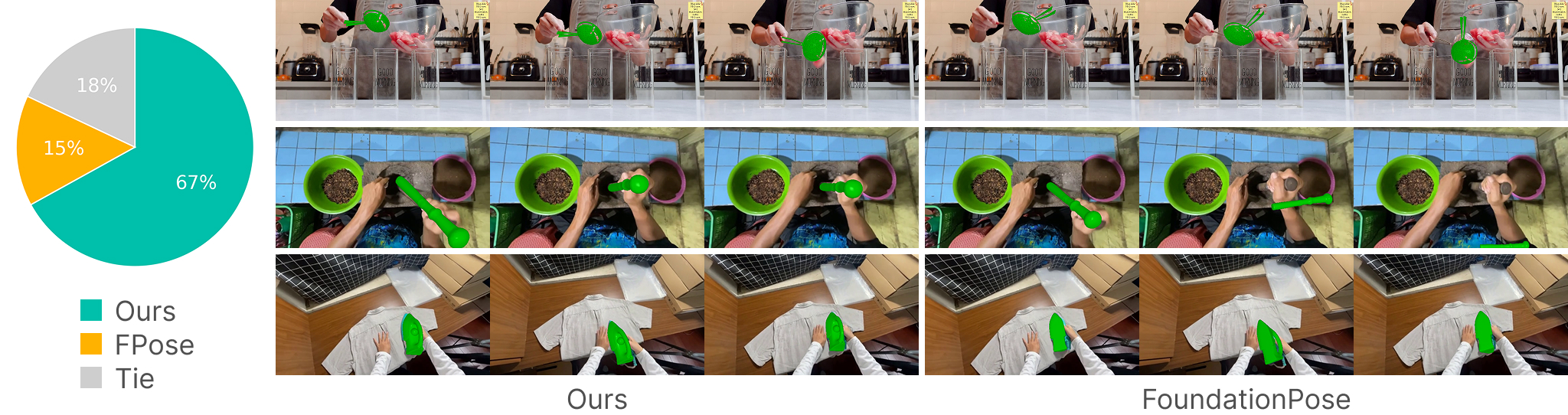}
\vspace{-1em}
\caption{\textbf{Object Tracking Comparison.} We compare Ours and FoundationPose~\citep{wen2024foundationposeunified6dpose} for object tracking with head-to-head human evaluations on 150 videos (left), and visualize samples (right).}
\label{fig:tracking-comparison}
\vspace{-0.5em}
\end{figure}

As shown in~\cref{fig:tracking-comparison} (left), on the 150-video in-the-wild benchmark, human raters prefer our object tracking over the state-of-the-art FPose 67\% of the time, with most videos receiving unanimous preferences.
Qualitatively (\cref{fig:tracking-comparison}), FPose loses the object under mild motion blur and occlusion, whereas our method recovers temporally consistent translations and rotations across the full clip. This advantage carries over to the standard hand-object reconstruction benchmarks in~\cref{tab:reconstruction}: we establish a new state-of-the-art on both DexYCB and HOI4D, outperforming all baselines.

\begin{table}[t]
\centering
\caption{\textbf{Reconstruction Results.}  F-5, F-10, and Chamfer distance (CD) on two hand-object tracking datasets.}
\label{tab:reconstruction}
\begin{tabular}{l ccc | ccc}
\toprule
& \multicolumn{3}{c|}{\textbf{DexYCB}} & \multicolumn{3}{c}{\textbf{HOI4D}} \\
\cmidrule(lr){2-4} \cmidrule(lr){5-7}
& F-5\,$\uparrow$ & F-10\,$\uparrow$ & CD\,$\downarrow$
& F-5\,$\uparrow$ & F-10\,$\uparrow$ & CD\,$\downarrow$ \\
\midrule
HO~\citep{hasson2019learning}    & 0.24 & 0.48 & 4.76 & 0.28 & 0.51 & 3.86 \\
IHOI~\citep{ye2022s}  & -- & -- & -- & 0.42 & 0.70 & 2.7 \\
HORSE~\citep{prakash20243d} & 0.23 & 0.42 & 6.97 & 0.26 & 0.45 & 6.69 \\
MCC-HO~\citep{wu2024reconstructing}  & 0.36 & 0.60 & 3.74 & 0.52 & 0.78 & 1.36 \\
G-HOP~\citep{ye2023ghop} & 0.31 & 0.49 & 8.11 & 0.69 & \textbf{0.91} & 0.63 \\
\midrule
FoundationPose~\citep{wen2024foundationposeunified6dpose} & 0.69 &  0.89  & 0.89 & 0.71 & \textbf{0.91} &  \textbf{0.49} \\
Any6D~\citep{lee2025any6d} & 0.69 & 0.88 & 0.97 & 0.71 & \textbf{0.91} & 0.50 \\
\midrule
\textbf{Ours} & \textbf{0.71} &  \textbf{0.93}  &  \textbf{0.66} & \textbf{0.72} & \textbf{0.91} & \textbf{0.49} \\
\bottomrule
\end{tabular}
\end{table}

In addition, we ablate design choices for our object tracking in the Appendix. We see that adaptive pose guidance via point tracking consistently improves reconstruction quality, and clustering-based selection performs on par with pose-likelihood selection while being up to 30$\times$ faster.

\subsection{Retargeting Results}

\begin{table}[t]
    \centering
    \caption{\textbf{Retargeting Results.} Success rate and average position and orientation error on two human-object reference datasets.}
    \label{tab:retargeting}
    \begin{tabular}{l ccc | ccc}
        \toprule
        & \multicolumn{3}{c|}{\textbf{Reconstruction}} & \multicolumn{3}{c}{\textbf{OakInk2}} \\
        \cmidrule(lr){2-4} \cmidrule(lr){5-7}
        Method & Success $\uparrow$ & Pos $\downarrow$ & Rot $\downarrow$ & Success $\uparrow$ & Pos $\downarrow$ & Rot $\downarrow$ \\
        \midrule
        Annealed Sampling             & 0.25 & 0.08 & 0.40 & 0.72 & 0.08 & 0.32 \\
        + Warmup             & 0.66 & 0.06 & \textbf{0.28} & 0.77 & 0.06 & 0.25 \\
        + Perturbation            & 0.67 & 0.06 & 0.30 & 0.79 & \textbf{0.03} & \textbf{0.14} \\
        + Transition Reward  & \textbf{0.71} & \textbf{0.05} & \textbf{0.28} & \textbf{0.81} & \textbf{0.03} & 0.15 \\
        \bottomrule
    \end{tabular}
\end{table}

We report retargeting results in Table~\ref{tab:retargeting}. On our reconstructed in-the-wild data, \method{} reaches a 71\% success rate, significantly improving over the baseline of 25\%. The main differentiator is warmup, which discovers initial states that are much more stable and natural than the noisy initial frame, thereby leading to successful tracking in subsequent timesteps. Additionally, we find that perturbation noticeably improves the qualitative results (e.g., natural grasps) despite marginally affecting the quantitative metrics, and our transition reward encourages successful picks and places for trajectories that otherwise would've missed the object during crucial transition timesteps.

Further validating our method on OakInk2, we also see consistent improvement with the introduction of each component, moving from a baseline of 72\% up to 81\%. This shows that our retargeting, despite being designed for imperfect reconstructed references, produces effective gains even with clean MoCap trajectories, and scales well to the 1,000+ bimanual tasks in this benchmark.

\subsection{Real-World Deployment}
\vspace{-0.5em}
\begin{figure}
\centering
\includegraphics[width=\columnwidth]{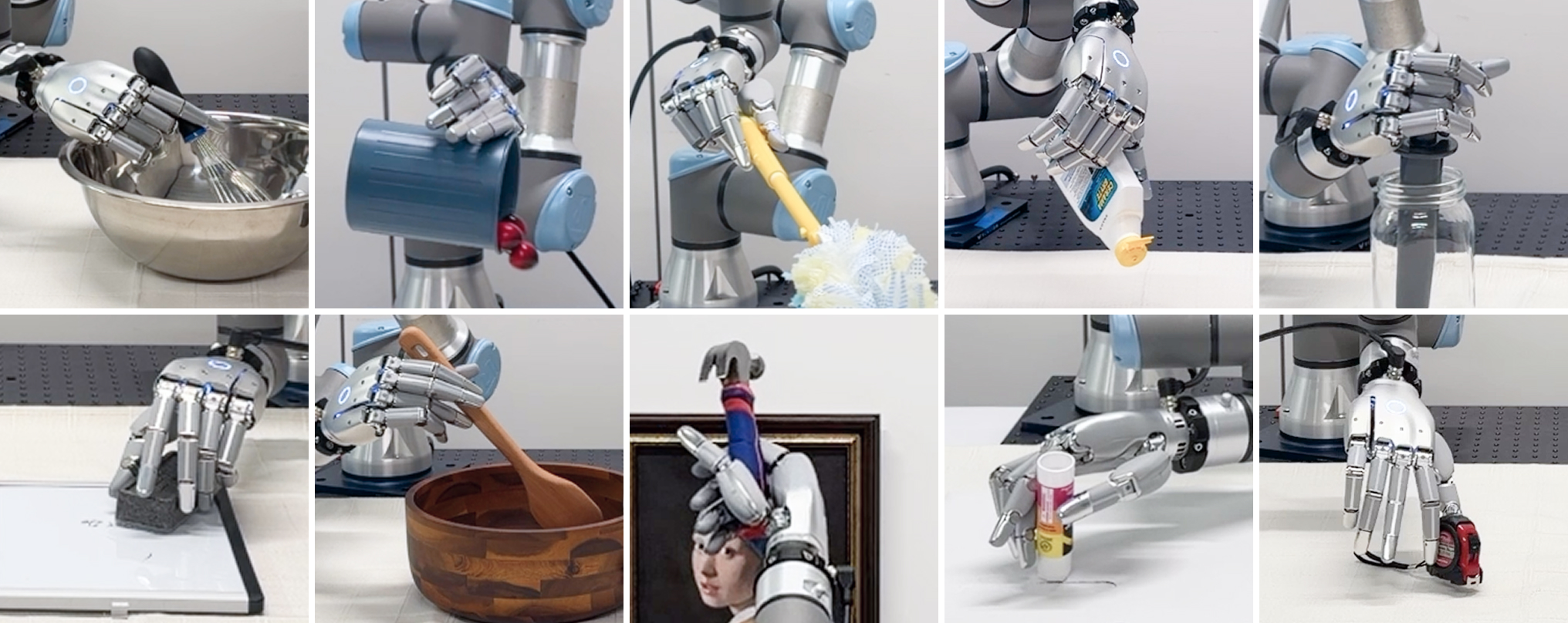}
\vspace{-1.0em}
\caption{\textbf{Real-World Deployment.} We showcase trajectories for 10 tasks: whisking, pouring, dusting, squeezing, tamping, erasing, stirring, hammering, spreading, and picking.}
\label{fig:robot}
\vspace{-0.5em}
\end{figure}

In total, our pipeline produced 500 high-quality, human-verified dexterous manipulation trajectories across internet (53\%), egocentric (31\%), and generated (16\%) videos. To demonstrate quality, we execute a representative set of trajectories in the real world. We choose 10 motions with various object geometries and grasp classes~\citep{feix2015grasp}, including writing tripod, power, ventral, and parallel extension grasps. After gravity alignment, the reconstructions still follow the videos' camera coordinates, so we manually align the initial pose ($x,y,z$, yaw) with the robot workspace in simulation before computing arm IK and deploying in the real world. Results are shown in~\cref{fig:robot}, further film strips are presented in the Appendix, and videos are presented on our \href{https://do-as-i-do.com/}{webpage}.
\vspace{-0.5em}
\subsection{Human Data Filtering Playbook}\vspace{-0.5em}
Given recent interest in human data for scaling up robotics~\cite{zheng2026egoscalescalingdexterousmanipulation, punamiya2026egoverseegocentrichumandataset,hoque2025egodex}, this section aims to highlight common quality issues in online human data sources. We noticed these patterns while analyzing common datasets, and present an analysis performed on 100DOH~\citep{shan2020understanding}. However, these lessons should be applicable more generally. 
We start with 2,000 10-second clips sampled from 100DOH, which has already been filtered for hand-object interaction, and find that only 187 clips (9\%) have meaningful hand-object interaction present. Out of these 187 candidate clips, 41 clips have the hand or object outside the video boundary, and 29 clips have no activity or activity that spans across shot boundaries. Further 14 clips fail due to camera motion, and another 10 clips fail because of SAM\,3D, both of which may be fixed in the future with better models. We lose another 10 clips for other reasons. Out of the 2,000 videos sampled from 100DOH, only 83 (4\%) survive our quality check for the reconstruction pass. Even in the best case, we foresee 107 clips, or roughly 5\% of the data, being directly relevant for learning dexterous manipulation, implying a $20\times$ penalty in not properly preprocessing and filtering internet videos for robot learning.
\vspace{-0.25em}
\section{Conclusion}
\label{sec:conclusion}
\vspace{-0.25em}
We introduced \method{}, a framework for reconstructing and retargeting everyday human videos onto dexterous robot hands. Our method is effective across egocentric, exocentric, and online video sources, showing a path towards scaling robot data by simply observing humans.
We hope~\method{} pushes us closer to making human videos first-class citizens in the robot learning data landscape.

\textbf{Limitations.} Our approach assumes rigid objects and semi-accurate metric depth predictions from monocular RGB, and may fail when either assumption doesn't hold. Monocular observations also suffer from ambiguity in the true hand-object distance, making it difficult to distinguish physical contact from mere visual occlusion. In addition, our method reconstructs only the hand and an object, rather than the full scene. As a result, it cannot reason about environmental constraints such as obstacles or articulations. Such scene-level reasoning remains important even with perfectly accurate references, since human intention is expressed through not only hand-object but also hand-scene interactions.
Finally, the current physics simulators model the real world dynamics only approximately, which places an upper bound on the achievable real-world performance of our framework.


\acknowledgments{We thank \href{https://kyutai.org/}{Kyutai} for providing us with the compute resources for this project. We are grateful to Chaoyi Pan for guidance and insightful discussions on retargeting. We also thank Jane Wu and Hongsuk Choi for helpful advice and discussions on hand-object reconstruction. This work was supported by ONR MURI N00014-21-1-280. Haritheja Etukuru and William Liang are supported by the NSF Graduate Research Fellowship Program under Grant DGE 2146752. Pieter Abbeel holds concurrent appointments as a Professor at UC Berkeley and as an Amazon Scholar at Amazon. This paper describes work performed at UC Berkeley and is not associated with Amazon.}


\bibliography{example}  

\newpage
\appendix
\section{Reconstruction}


\begin{figure}[H]
\centering
\includegraphics[width=0.9\columnwidth]{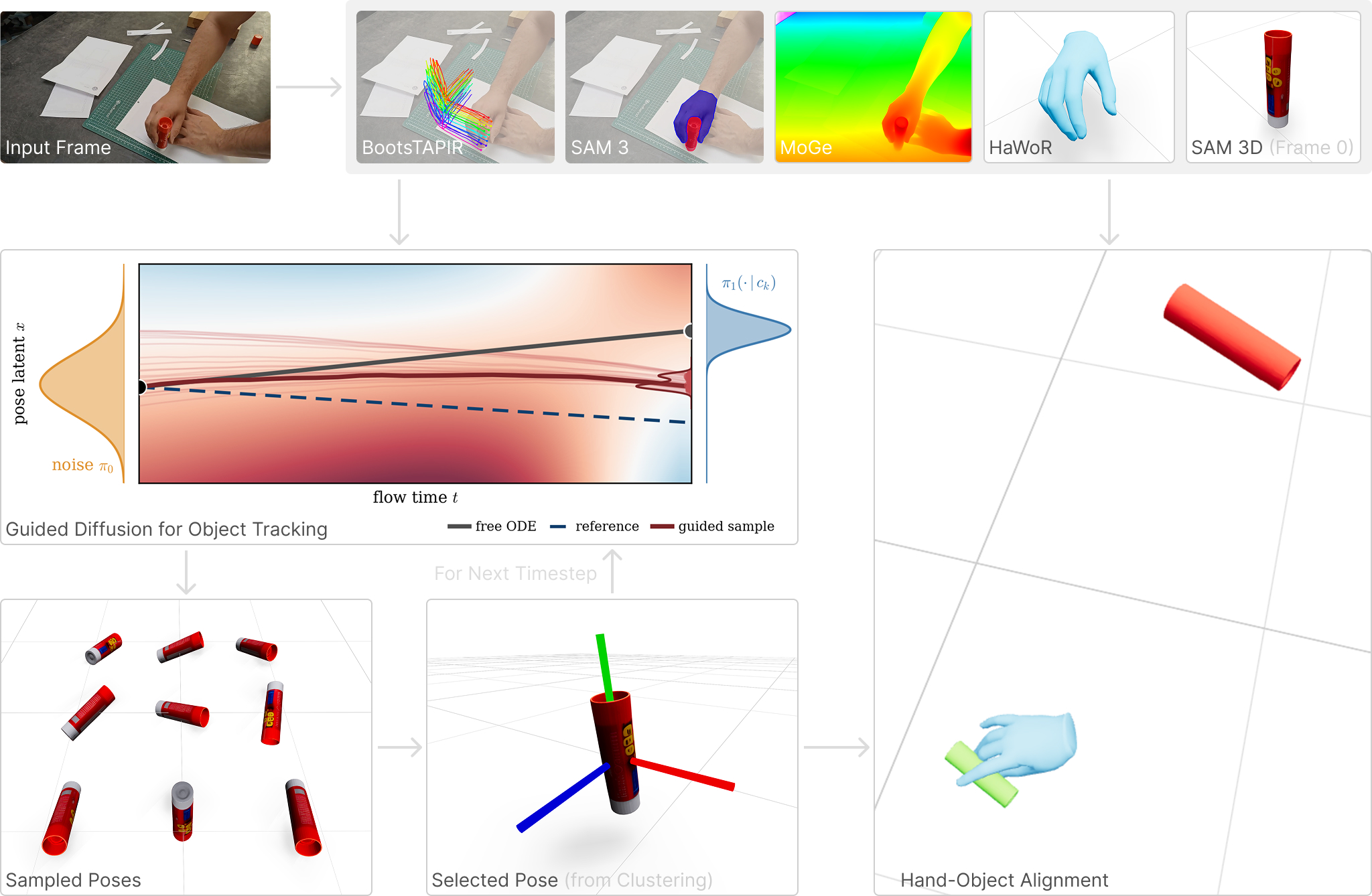}
\caption{\textbf{Reconstruction Architecture.} SAM 3D~\citep{sam3dteam2025sam3d3dfyimages} generates the object mesh from a single frame, while HaWoR \citep{zhang2025hawor} tracks the hand across the video. We then track the object frame-by-frame via guided diffusion  (Section 3.1), anchoring each step to the predicted object shape and the previous frame's pose. Per frame, we sample $N$ candidate poses and select the best using a clustering-based heuristic. Finally, a depth-map-based alignment step registers the HaWoR hand to our predicted object, yielding a consistent 4D hand-object trajectory.}
\label{fig:reconstruction}
\end{figure}

\textbf{Adaptive Guidance.}
Here, we explain how we get the pose guidance strength parameter $\alpha_p$ using point tracks. For each pair of consecutive frames, we sample $20$ points inside the object mask and track them with BootsTAPIR~\citep{doersch2024bootstapbootstrappedtrainingtrackinganypoint}. We retain the points that remain inside the next mask and fit a 2D rigid transform to them in closed form by SVD.\@ At every frame $k$, this yields the object's estimated in-plane rotation $\Delta\theta_{k}$ together with its centroid translation. We then set the pose guidance strength $\guide{\alpha_{p}}$ as a clamped affine function of the rotation magnitude. In general, we use the following function to estimate $\alpha_{p}(k)$:
$\alpha_{p}(k)=\max\!\left(0.1,\; 0.7-0.09\,\lvert\Delta\theta_{k}\rvert\right)$. For standard benchmarks we evaluate on DexYCB and HOI4D~\citep{chao2021dexycb, liu2022hoi4d}, and show that adaptive pose guidance consistently improves over fixed pose guidance experiments in Table~\ref{tab:reconstruction_ablations}.

\textbf{Sampling Per-frame Poses.} At each frame $k$, our algorithm draws $N$ candidates $\{x^{p}_{k,i}\}_{i=1}^{N}$ that share the fixed shape $\bar x^{s}$ and needs to pick one of the samples per frame. The principled choice is to rank candidates by the model's own conditional log-density over poses given a shape. 
For a flow model this density is exactly computable through the instantaneous change-of-variables formula~\citep{lipman2023flowmatchinggenerativemodeling}, restricted to the pose block:

\begin{equation}
  \log p_{\theta}\!\left(x^{p}_{k,i} \,\middle|\, \bar x^{s},\, c_{k}\right)
  \;=\;
  \log p_{0}\!\left(x^{p}_{0}\right)
  \;+\;
  \int_{0}^{1} \operatorname{tr}\!\left(
  \frac{\partial v^{p}_{\theta}(x_{t}, t, c_{k})}{\partial x^{p}_{t}}
  \right)\, dt,
  \label{eq:loglik}
\end{equation}
where $x^{p}_{0}$ is the noise pre-image of $x^{p}_{k,i}$ obtained by
integrating the ODE backward from $t{=}1$ to $t{=}0$. Evaluating the trace
exactly requires one vector-Jacobian product per pose coordinate at every
Euler step; with $D_{p}{=}13$ pose dimensions, $T{=}25$ ODE steps, and
$N{=}25$ candidates per frame, scoring costs
$N\,T\,(1{+}D_{p}) \approx 8.7\mathrm{k}$ forward+backward passes through the
diffusion backbone \emph{per frame} --- roughly two orders of magnitude above
generation itself and prohibitive at video scale.

As a result, we go with a clustering based heuristic which is almost real-time once the candidates have been generated. We sample and cluster $N = 25$ poses under a weighted $\mathrm{SE}(3)$ distance described below:
\begin{equation}
  d\!\left(x^{p}_{i}, x^{p}_{j}\right) \;=\;
  w_{t}\,\|t_{i} - t_{j}\|_{2}
  \;+\;
  w_{r}\cdot 2\arccos\bigl|\langle q_{i}, q_{j}\rangle\bigr|,
  \label{eq:posedist}
\end{equation}
where $(t_{i}, q_{i}):=x_i^p$ are the translation and unit-quaternion components of $x^{p}_{i}$ and the second term is the geodesic angle on $\mathrm{SO}(3)$. We discard clusters below a minimum size as outliers and rank the remaining clusters' 2D silhouette IoU against the input mask. 
Empirically, confident samples concentrate on the same pose mode while estimator noise scatters across $\mathrm{SE}(3)$. We show that this heuristic performs on part with pose likelihood based ranking in Table~\ref{tab:reconstruction_ablations} while adding essentially zero computational cost over generation.

\begin{figure}[H]
\centering
\includegraphics[width=\columnwidth]{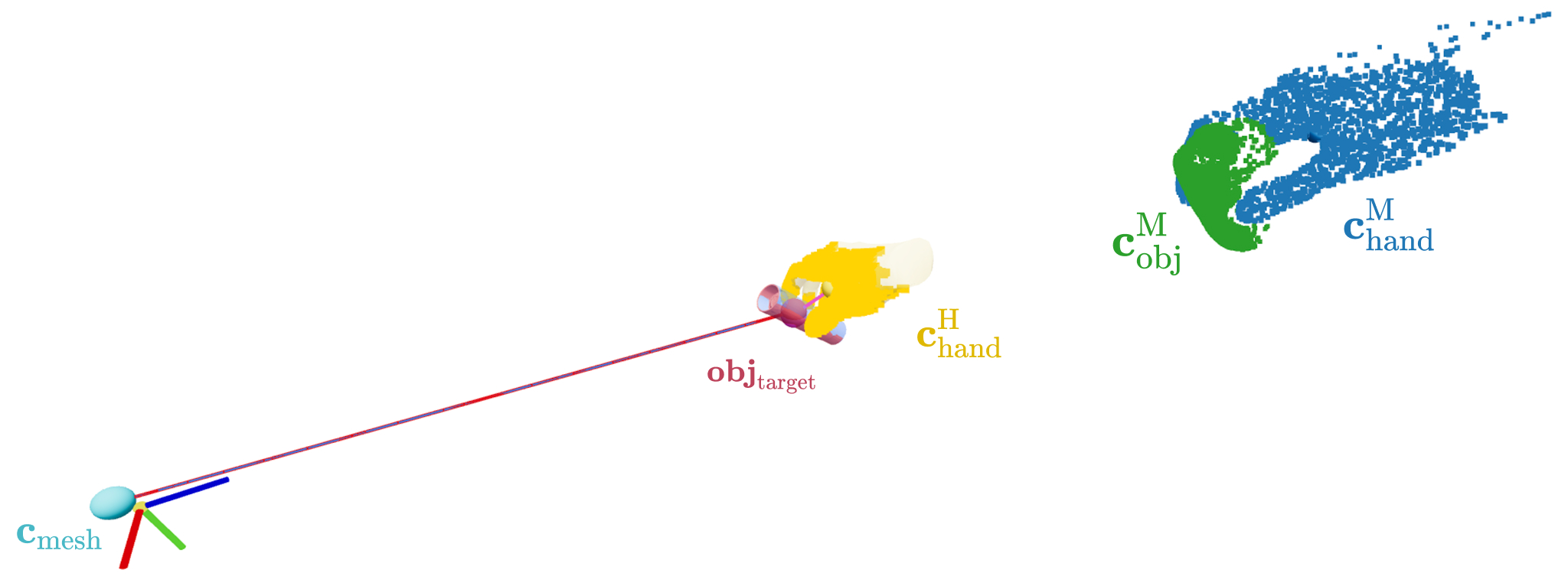}
\caption{\textbf{Hand-Object Alignment.} The translation and scale of the object mesh are converted from MoGe pointmap space to HaWoR hand mesh space using relative distance between hand and object.}
\label{fig:alignment}
\end{figure}

\textbf{Hand-Object Alignment.}
Since SAM 3D has been trained using pointmaps obtained from monocular geometry estimation model MoGe~\citep{wang2025moge2accuratemonoculargeometry}, we use MoGe pointmaps during inference as well for best performance. To align the pose predictions with the HaWoR hand predictions, as seen in Figure~\ref{fig:alignment}, we first compute the centroid of the \textit{visible} portion of the HaWoR hand mesh $\mathbf{c}^{\mathrm{H}}_{\mathrm{hand}}$ by casting a ray for each pixel in the segmentation mask, and then averaging over the first hits onto the hand mesh. Then, in the per-frame MoGe pointmap, the hand centroid $\mathbf{c}^{\mathrm{M}}_{\mathrm{hand}}$ is the mean over the $3$D values of the same pixels used to recover $\mathbf{c}^{\mathrm{H}}_{\mathrm{hand}}$ (the rays that strike the HaWoR mesh), and the object centroid $\mathbf{c}^{\mathrm{M}}_{\mathrm{obj}}$ is the mean of the $3$D values over all segmentation mask pixels. 
Letting $z^{\mathrm{H}}_{\mathrm{hand}}$ and $z^{\mathrm{M}}_{\mathrm{hand}}$ denote the depth ($z$) components of the hand centroids, the per-frame scale is\[k = \frac{z^{\mathrm{H}}_{\mathrm{hand}}}{z^{\mathrm{M}}_{\mathrm{hand}}}.\] For each frame we place the object relative to the hand: we take the hand-to-object offset in pointmap space, rescale it to the near-metric (HaWoR) units by $k$, and add it to this HaWoR hand centroid,
  \[
    \mathbf{obj}_{\mathrm{target}}
      = \mathbf{c}^{\mathrm{H}}_{\mathrm{hand}}
      + k \,\bigl(\mathbf{c}^{\mathrm{M}}_{\mathrm{obj}} - \mathbf{c}^{\mathrm{M}}_{\mathrm{hand}}\bigr).
  \]
Given this target object position, we hold the SAM 3D predicted mesh orientation fixed and optimize a single scalar, the translation scale $s$, on the mesh's camera-frame translation $\mathbf{t}$. The object's visible-surface centroid is $\mathbf{obj}_{\mathrm{pos}}(s) = \mathbf{c}_{\mathrm{mesh}} + s\,\mathbf{t}$, where $\mathbf{c}_{\mathrm{mesh}}$ is the centroid of \textit{visible} mesh vertices that project inside the object mask (with translation excluded). We solve the one-dimensional least-squares problem
  \[
    s^{\star}
      = \arg\min_{s}\,\bigl\| \mathbf{obj}_{\mathrm{pos}}(s) - \mathbf{obj}_{\mathrm{target}} \bigr\|^{2}
      = \frac{\mathbf{t}^{\top}\bigl(\mathbf{obj}_{\mathrm{target}} - \mathbf{c}_{\mathrm{mesh}}\bigr)}
             {\mathbf{t}^{\top}\mathbf{t}},
  \]
which slides the object along its viewing ray to the depth that best matches the target while preserving its recovered orientation. Repeating this for every frame yields an aligned metric 4D hand-object trajectory. 

\section{Retargeting}

\textbf{Simulation Setup.} We perform retargeting in the MuJoCo Warp simulator, with a simulation timestep duration of 0.005s (200 Hz). We convex decompose object meshes using CoACD~\citep{Wei_2022}, and to stabilize hand-object interactions especially for tasks that involve many contacts, we thicken and dilate the object meshes by 2mm.

\begin{figure}[H]
\centering
\includegraphics[width=\columnwidth]{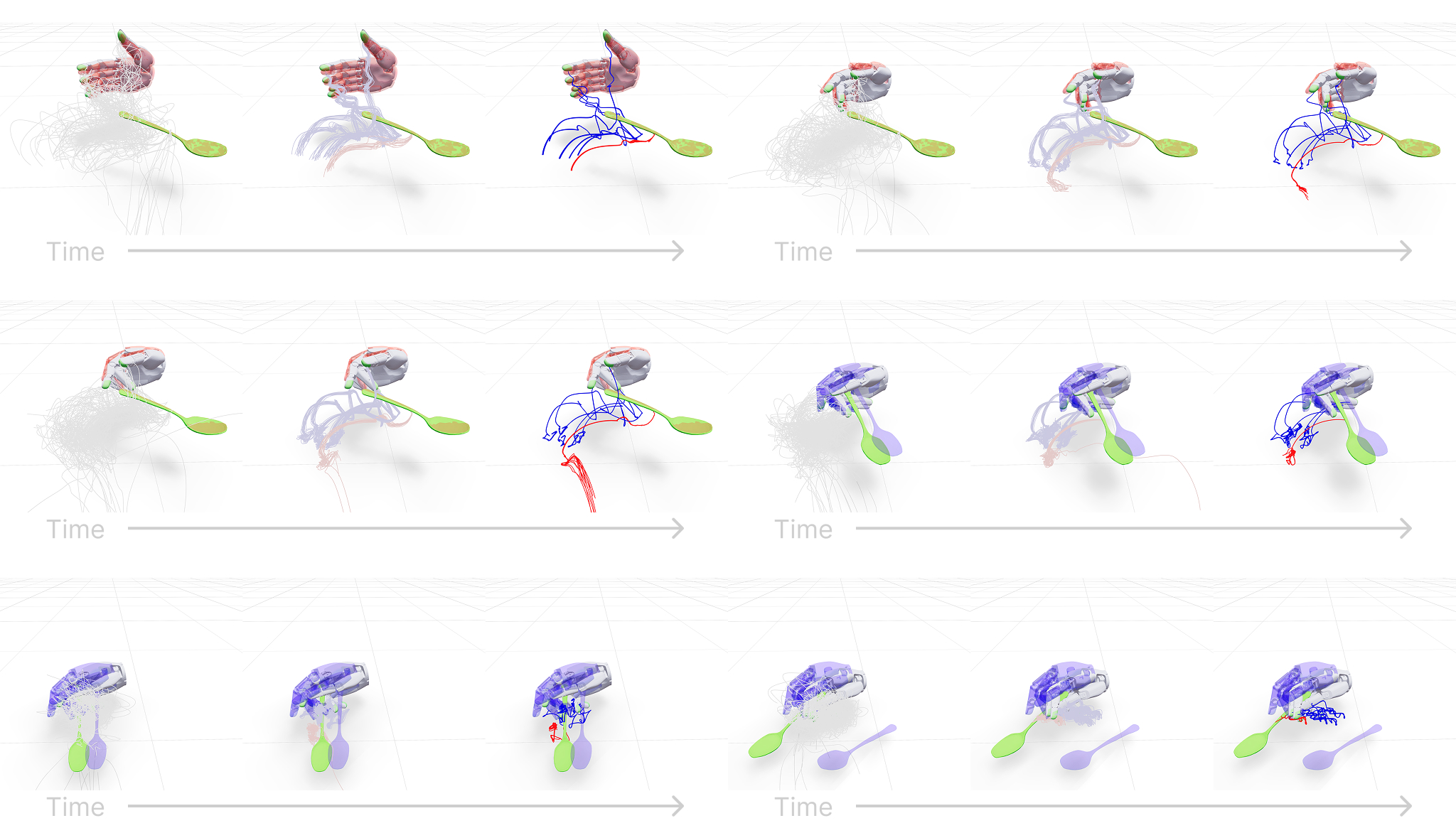}
\caption{\textbf{Retargeting Optimization.} We visualize multiple iterations of the sampling-based optimization process for a trajectory: blue and red traces indicate converged fingertip and object trajectories, respectively. The ghost hand and object indicate reference (blue) and warmup (red).}
\label{fig:retargeting-sampling}
\end{figure}

\textbf{Algorithm Details.} To prepare for dynamics-aware retargeting, we first compute the reference trajectory by kinematically retargeting the human hand onto the robot hand (without considering forces, contacts, etc), using mink to match fingertip positions~\citep{Zakka_Mink_Python_inverse_2026}. Next, we run our sampling-based dynamics-aware retargeting algorithm, visualized in~\cref{fig:retargeting-sampling}, with planning every 0.5s (2 Hz) on a horizon of 3s. Each planning step, we evaluate 1024 samples and optimize over 32 iterations, and rollouts are rewarded for tracking the object (position and orientation) and hand (position, orientation, and finger joints), with penalties on excessive penetration (to avoid exploiting the simulator) and the transition reward introduced in the main text. Further hyperparameter details are in~\cref{tab:retargeting-setup}.

\begin{table}[h]
\centering
\caption{\textbf{Retargeting Hyperparameters.} All results from our method use the following setup.}
\label{tab:retargeting-setup}
\begin{tabular}{l l | l l}
\toprule
\textbf{Parameter} & \textbf{Value} & \textbf{Parameter} & \textbf{Value} \\
\midrule
\multicolumn{2}{l}{\textit{Algorithm}} & \multicolumn{2}{l}{\textit{Sampling}} \\
\midrule
\texttt{num\_samples}         & 1024  & \texttt{knot\_dt}                  & 0.2  \\
\texttt{max\_num\_iterations} & 32    & \texttt{pos\_noise\_scale}         & 0.01 \\
\texttt{sim\_dt}              & 0.005 & \texttt{rot\_noise\_scale}         & 0.01 \\
\texttt{ctrl\_dt}             & 0.5   & \texttt{joint\_noise\_scale}       & 0.1  \\
\texttt{horizon}              & 3.0   & \texttt{final\_noise\_scale}       & 0.01 \\
                              &       & \texttt{first\_ctrl\_noise\_scale} & 1.0  \\
                              &       & \texttt{last\_ctrl\_noise\_scale}  & 4.0  \\
\midrule
\multicolumn{2}{l}{\textit{Rewards}} & \multicolumn{2}{l}{\textit{Perturbation}} \\
\midrule
\texttt{pos\_rew\_scale}             & 1.0    & \texttt{num\_perturb\_samples}   & 4    \\
\texttt{rot\_rew\_scale}             & 0.3    & \texttt{perturb\_force\_scale}   & 0.5  \\
\texttt{base\_pos\_rew\_scale}       & 0.1    & \texttt{perturb\_torque\_scale}  & 0.5  \\
\texttt{base\_rot\_rew\_scale}       & 0.03   & \texttt{perturb\_prob}           & 0.05 \\
\texttt{joint\_rew\_scale}           & 0.01   & \texttt{perturb\_continue\_prob} & 0.95 \\
\texttt{terminal\_rew\_scale}        & 10.0   &                                  &      \\
\texttt{penetration\_penalty\_scale} & 3000.0 &                                  &      \\
\texttt{transition\_penalty\_scale}    & 0.5    &                                  &      \\
\bottomrule
\end{tabular}
\end{table}

Finally, we list several important details and lessons besides those described in the main text.
\begin{enumerate}
    \item \textbf{Reference Blending.} At each planning timestep, the initial samples are centered around the previous plan's controls, appended at the end with the next chunk of reference steps. Naively appending the reference to previously optimized controls results in sharp transitions and jerky motions, and to avoid this, we blend the optimized controls into the reference via interpolation.
    \item \textbf{Robust Kinematic Retargeting.} Retargeting draws samples centered around the reference, and thus requires a reasonable reference in order to find dynamically-feasible controls. Since kinematic retargeting is computationally inexpensive, a simple yet effective way to improve reference quality is to compute multiple kinematic retargeting results starting from different random initial poses, thus avoiding potential local minima.
    \item \textbf{Object Base.} In some videos, objects of interest may not be lying flat on a surface (e.g., spoon standing upright in a container). To faithfully stabilize their initial poses, we add a flat base ``plate'' to the bottom of the object mesh, which only has contact with the floor (and not the robot). This allows us to retarget objects starting from any initial pose, without any task-specific assumptions or additional scene reconstructions.
\end{enumerate}

\section{Experimental Setup and Results}

\textbf{Reconstruction Evaluation.} We compare against two groups of baselines: (1) joint hand-object reconstructions, and (2) object trackers~\citep{wen2024foundationposeunified6dpose, lee2025any6d}. In the first group, we compare against HO~\citep{hasson2019learning}, IHOI~\citep{ye2022s}, HORSE~\citep{prakash20243d}, and MCC-HO~\citep{wu2024reconstructing}, whose numbers are taken from~\citet{wu2024reconstructing}. We also compare against a more recent video-based approach for RGB joint hand-object reconstruction G-HOP~\citep{ye2023ghop}. For HOI4D, we evaluate the authors' released per-clip checkpoints directly. These checkpoints were optimized with the HOI4D-only diffusion prior provided. For DexYCB, the original GHOP paper trains the diffusion prior on (among other datasets) the DexYCB shape and pose data, but does not report video-reconstruction results on DexYCB. We therefore run the per-clip test-time optimization ourselves on the full 160-clip test set using the authors' released mix-data prior. In the next group of baselines, we have state-of-the-art 6-DoF object trackers, FoundationPose~\citep{wen2024foundationposeunified6dpose} and Any6D~\citep{lee2025any6d}. Since the latter methods only perform object tracking, we slot each into our framework in place of our object-tracking module while holding every other component fixed, providing a controlled comparison against our tracker.

\begin{figure}[H]
\centering
\includegraphics[width=\columnwidth]{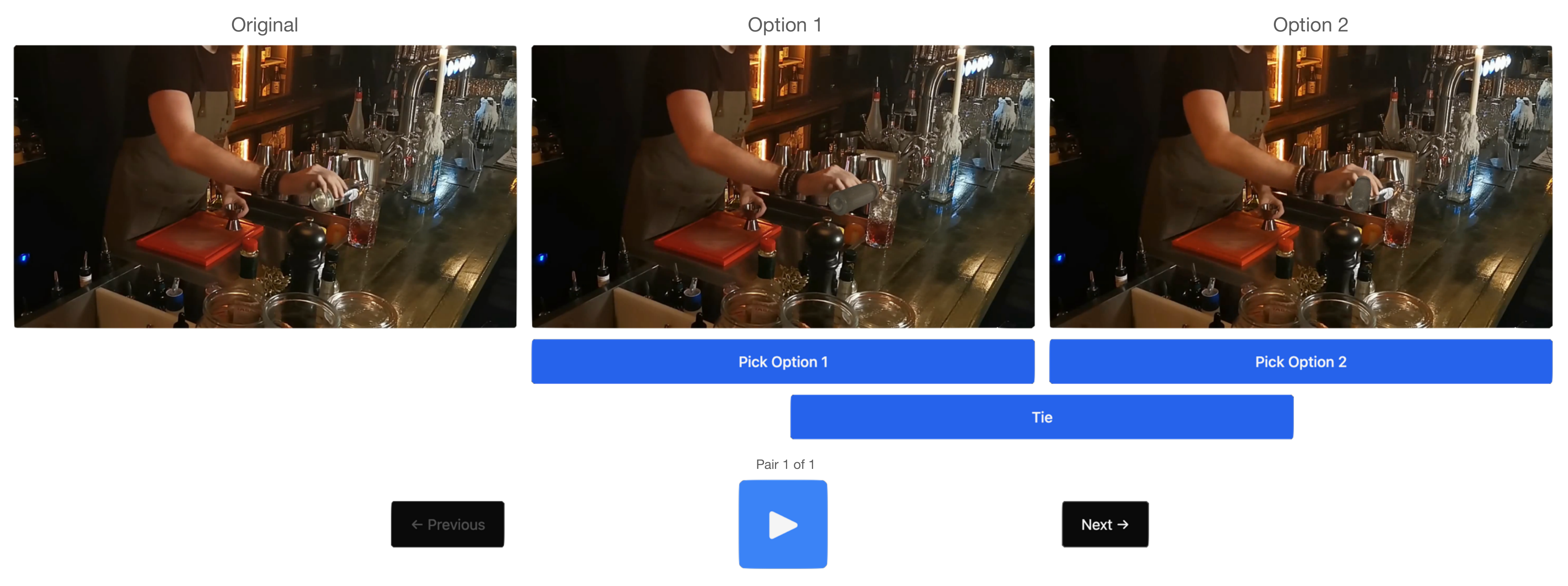}
\caption{A screenshot of the user interface shown to the human evaluators for in-the-wild object tracking. In this instance shown in the figure, option 1 is regarded as better because the object pose is aligned with true object position.}
\label{fig:eval_ui}
\end{figure}

For our evaluation on in-the-wild internet videos, ground-truth object poses are unavailable, so we rely on human preference. Each rater is shown the original hand-object interaction video alongside two reprojections, where the posed object mesh from our tracker and from FoundationPose \citep{wen2024foundationposeunified6dpose} is projected back into 2D, and asked which video tracks the object more consistently with its true motion. We randomize the left-right ordering of the two methods per video to avoid position bias, and collect three ratings, each from a distinct rater in a pool of five, for each of the 150 videos. Raters prefer our method 67\% of the time, compared to 18\% for FoundationPose and 15\% rated as ties. This corresponds to a 79\% win rate among non-tie judgments. 75\% of videos received unanimous agreement across all three raters, and inter-rater agreement was substantial (Fleiss' $\kappa = 0.65$). Figure~\ref{fig:eval_ui} shows a screenshot of the user interface presented to each rater.

\begin{table}[t]
\centering
\caption{\textbf{Object Tracking Ablation.} We ablate two design axes of our object tracking method.}
\label{tab:reconstruction_ablations}
\begin{tabular}{cc ccc ccc}
\toprule
\multirow{2}{*}{\shortstack{\textbf{Pose Guidance}\\ \textbf{Strategy}}}
& \multirow{2}{*}{\shortstack{\textbf{Candidate}\\ \textbf{Selection}}}
& \multicolumn{3}{c}{\textbf{DexYCB}} & \multicolumn{3}{c}{\textbf{HOI4D}} \\
\cmidrule(lr){3-5} \cmidrule(lr){6-8}
& 
& F-5\,$\uparrow$ & F-10\,$\uparrow$ & CD\,$\downarrow$
& F-5\,$\uparrow$ & F-10\,$\uparrow$ & CD\,$\downarrow$ \\
\midrule
 Fixed & Clustering & 0.70 & 0.91 & 0.74 & 0.69 & 0.91 & 0.50 \\

Adaptive & Random & 0.70 & 0.91 & 0.74 & 0.62 & 0.87 & 0.66 \\
Adaptive & Log-likelihood &   0.72    &  0.93    &   0.65 &  0.72    &  0.91    &   0.49   \\
Adaptive & Clustering & 0.71 & 0.93 & 0.66 & 0.72 & 0.91 & 0.49 \\

\bottomrule
\end{tabular}
\end{table}

\section{Robot Deployment}

\begin{figure}[H]
\centering
\includegraphics[width=0.95\columnwidth]{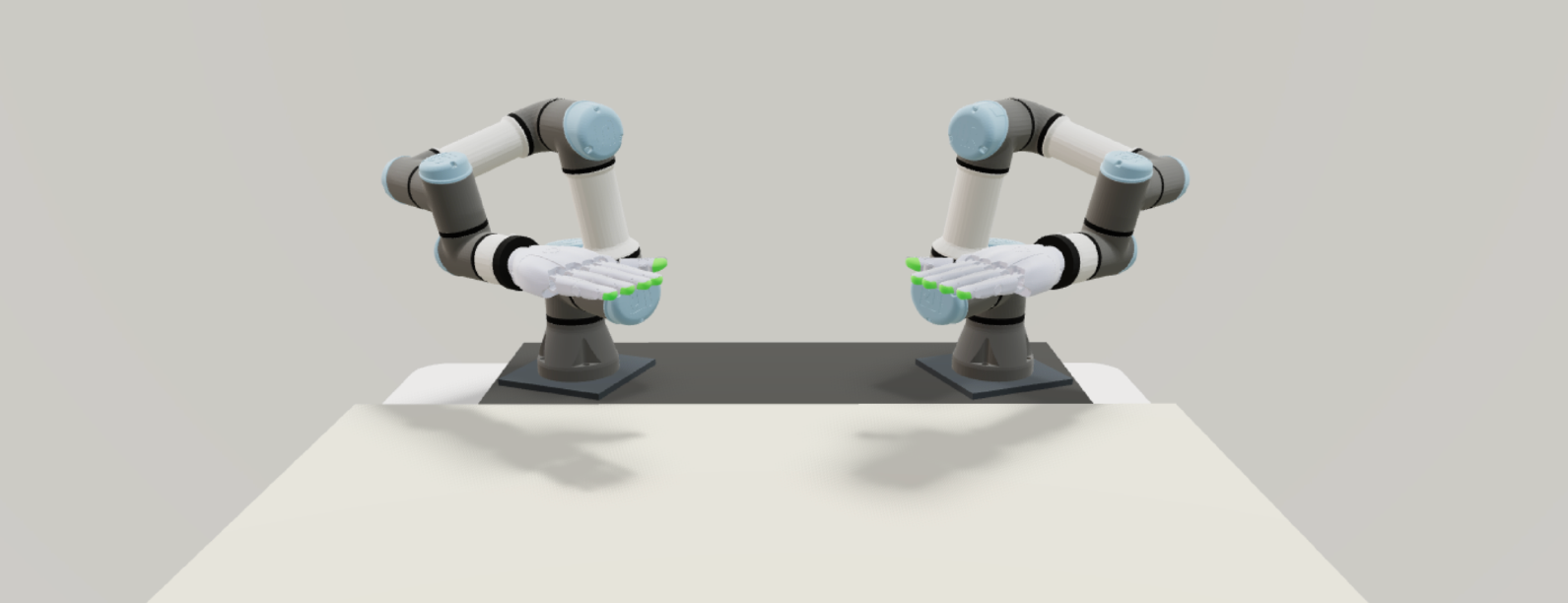}
\caption{\textbf{Digital Twin.} A simulated replica of our real-world bimanual setup (UR3e arms with Sharpa Wave hands) in MuJoCo, visualized with Viser. }
\label{fig:twin}
\end{figure}

\textbf{Simulation Setup for Dual UR3e arms + Sharpa hands.}
We replay the retargeted trajectories in simulation on a dual-arm setup with UR3e arms and Sharpa Wave hands, using the MuJoCo physics simulator~\citep{todorov2012mujoco} for dynamics and Viser for visualization. The trajectories from our retargeting stage (Section 3.2) cannot be executed on the robot directly; we first map them onto the arm-and-hand setup via inverse kinematics using mink \citep{Zakka_Mink_Python_inverse_2026}. As shown in Figure~\ref{fig:twin}, we build a digital twin of our real-world setup, letting us visually validate each trajectory for self-collisions, table contacts, and similar issues in simulation before real-world execution.

\textbf{Real-World Rollouts.}
Once validated in simulation, we roll out the trajectories on the real-world dual-arm setup at roughly half speed, with both arms and hands commanded at 50 Hz. Figure~\ref{fig:rollout} shows the real-world rollout strips for 6 different tasks:

\begin{figure}[H]
\centering
\includegraphics[width=\columnwidth]{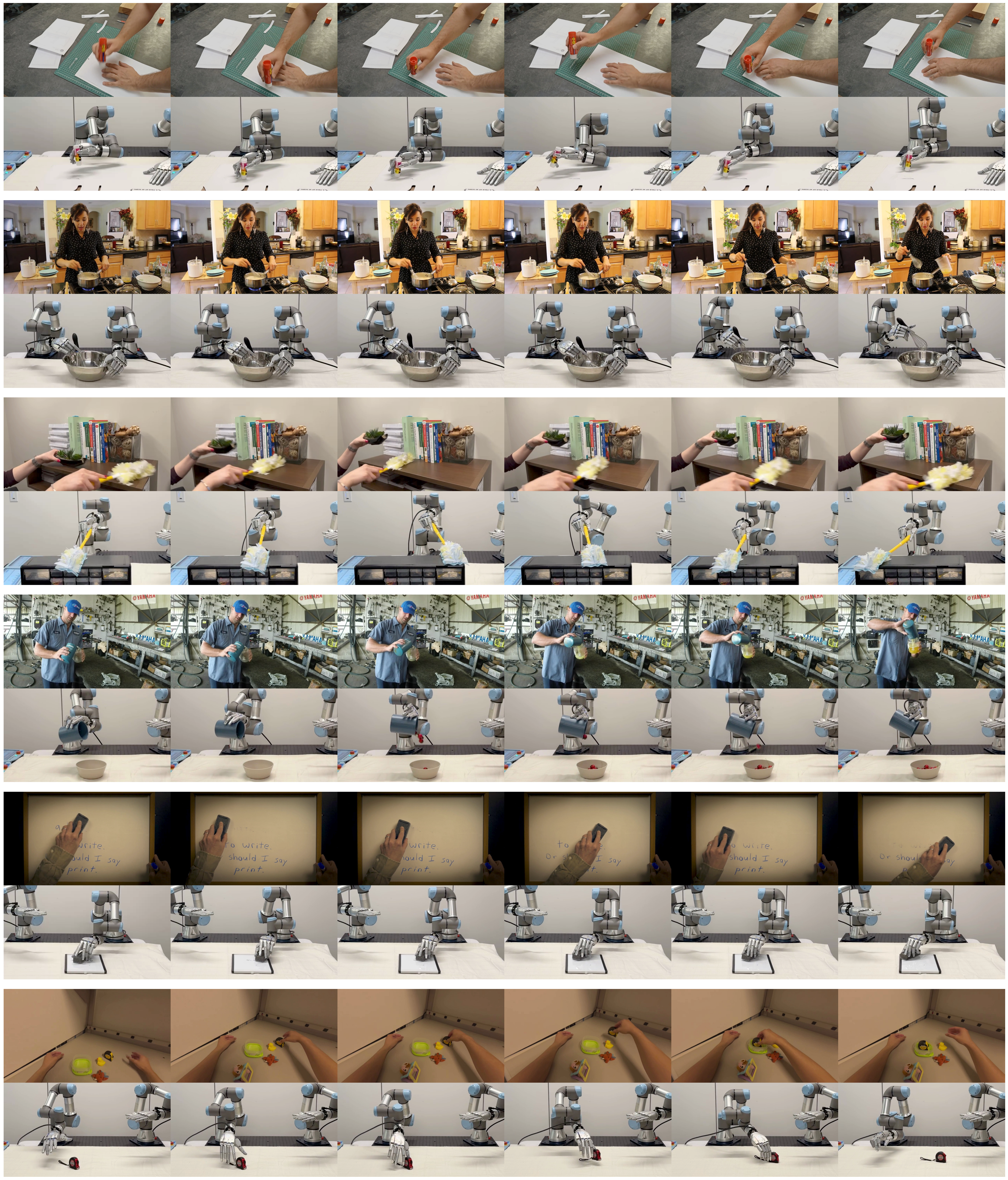}
\caption{\textbf{Real-World Rollouts.} Frames from our robot rollouts for spreading, whisking, dusting, pouring, erasing, and picking. More tasks and videos are available on our webpage.}
\label{fig:rollout}
\end{figure}



\end{document}